\documentclass[runningheads]{llncs}

\usepackage{accv}

\usepackage{accvabbrv}

\usepackage{graphicx}
\usepackage{booktabs}
\usepackage{multirow} 
\usepackage{amssymb}
\usepackage{listings}
\usepackage{colortbl}
\usepackage{pifont}

\usepackage{algorithm}
\usepackage{algpseudocode}
\usepackage{amsmath}
\definecolor{common_blue}{RGB}{66,115,116}

\usepackage[accsupp]{axessibility}  

\def\eg{\emph{e.g}\onedot} 
\def\ie{\emph{i.e}\onedot}

\def\etal{\emph{et al}\onedot}
\def\@onedot{\ifx\@let@token.\else.\null\fi\xspace}

\usepackage[pagebackref,breaklinks,colorlinks,citecolor=accvblue]{hyperref}
\hypersetup{
        final,
        colorlinks,
        linkcolor={red!60!black},
        citecolor={blue!60!black},
        urlcolor={blue!30!black}
        }

\usepackage{orcidlink}

\newcommand\blfootnote[1]{%
  \begingroup
  \renewcommand\thefootnote{}\footnote{#1}%
  \addtocounter{footnote}{-1}%
  \endgroup
}

\begin{document}

\title{Few Exemplar-Based \\General Medical Image Segmentation via Domain-Aware Selective Adaptation} 

\titlerunning{FEMed: few exemplar-based general medical image segmentation} 

\author{Chen Xu\inst{1,2}\textsuperscript{$\star$} \and
Qiming Huang\inst{1}\textsuperscript{$\star$} \and
Yuqi Hou\inst{1}\and
Jiangxing Wu \inst{2}\and
Fan Zhang \inst{2}\and
Hyung Jin Chang \inst{1}\and
Jianbo Jiao \inst{1}
}

\authorrunning{C.~Xu, Q.~Huang~\etal}
\institute{
School of Computer Science, University of Birmingham, Birmingham, UK \\
\and
School of Computer Science, Fudan University, Shanghai, China\\
Project page: \href{https://xuchenjune.github.io/FEMed/}{https://xuchenjune.github.io/FEMed/}
\blfootnote{\textsuperscript{$\star$} Equal contribution.}
}
\maketitle

\begin{abstract}

Medical image segmentation poses challenges due to domain gaps, data modality variations, and dependency on domain knowledge or experts, especially for low- and middle-income countries (LMICs). Whereas for humans, given a few exemplars (with corresponding labels), we are able to segment different medical images even without extensive domain-specific clinical training. In addition, current SAM-based medical segmentation models use fine-grained visual prompts, such as the bounding rectangle generated from manually annotated target segmentation mask, as the bounding box (bbox) prompt during the testing phase. However, in actual clinical scenarios, no such precise prior knowledge is available. Our experimental results also reveal that previous models nearly fail to predict when given coarser bbox prompts. Considering these issues, in this paper, we introduce a domain-aware selective adaptation approach to adapt the general knowledge learned from a large model trained with natural images to the corresponding medical domains/modalities, with access to only a few (\eg less than 5) exemplars. Our method mitigates the aforementioned limitations, providing an efficient and LMICs-friendly solution. Extensive experimental analysis showcases the effectiveness of our approach, offering potential advancements in healthcare diagnostics and clinical applications in LMICs.

\end{abstract}

\section{Introduction}
\label{sec:intro}

Medical image segmentation is a fundamental problem in healthcare and clinical applications. \textcolor{black}{Achieving high-quality segmentation of medical images in specific domains or modalities (\eg MRI, CT) typically requires the expertise of domain specialists} with extensive medical training in that particular data modality.
With the development of deep neural networks (DNNs) ~\cite{shin2016deep}, \textcolor{black}{it has become} possible to automate this segmentation process by training a DNN. To achieve this, the DNN model needs sufficient (usually large-scale) training data, \ie input image and the paired ground-truth segmentation masks. However, the collection of \textcolor{black}{such paired} data (especially the segmentation mask annotation) is time-consuming and costly, sometimes even infeasible to acquire. Such challenges have significantly hindered the development of automatic medical image segmentation.

On the other hand, for humans, even without domain-specific professional clinical training, given an example (or a few) of how a medical image was segmented, usually the person can segment similar images in the same medical domain. This is attributed to the segmentation-related prior knowledge plus \textcolor{black}{guidance from a few examples}. Usually, the more examples are exposed to, the better the person can perform in that medical domain.
Considering the above observations, we are interested in asking: \textit{is that possible to train an automatic medical segmentation model only with general prior knowledge and a few exemplars (rather than extensive domain expert annotations)?}

\begin{figure}[t]
\begin{center}
    \includegraphics[width=0.9\textwidth]{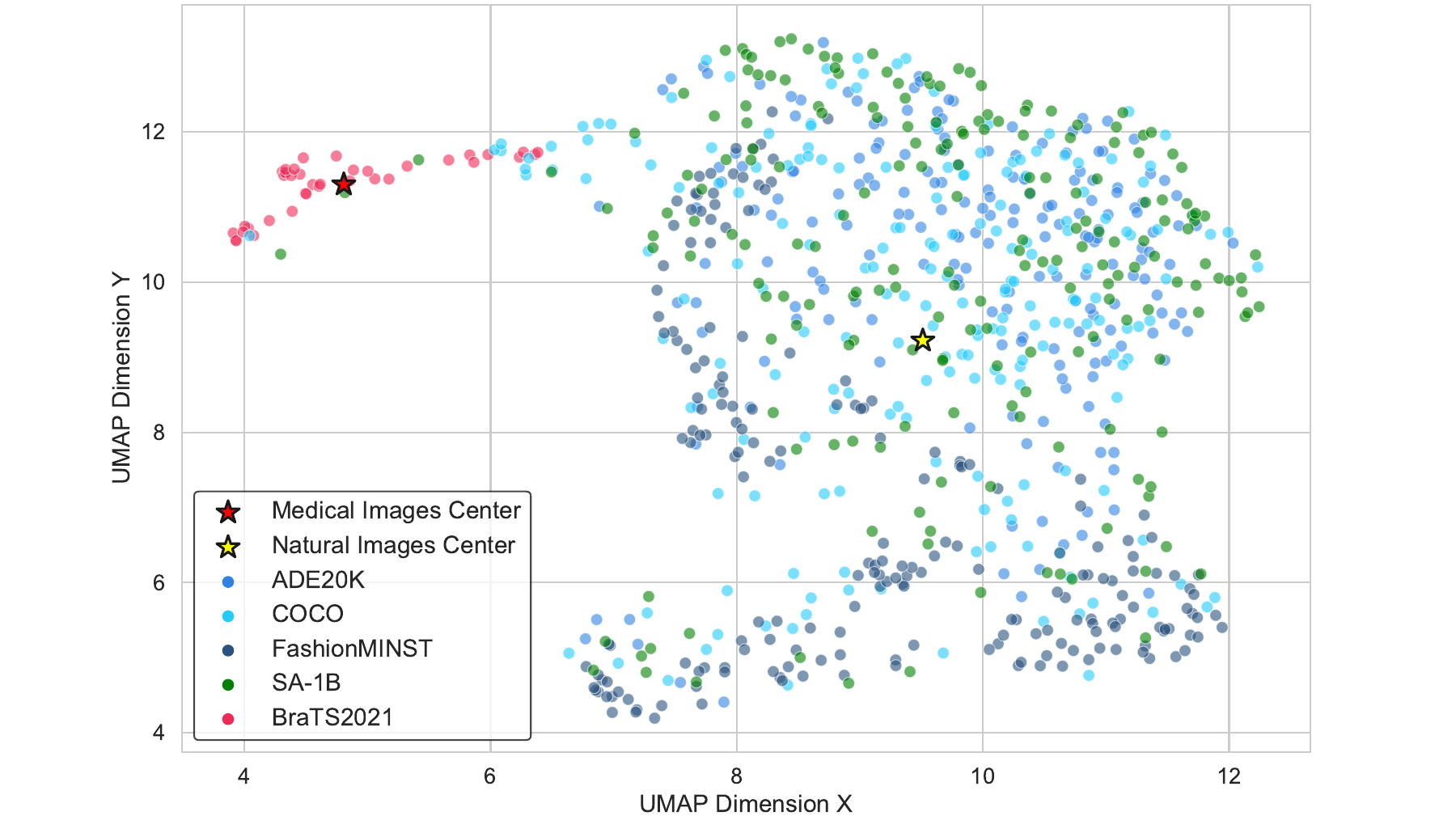}
\end{center}
\caption{Illustration of the relative relationship and distribution of samples from different datasets, using~\cite{SMG2020} for feature visualisation. A clear domain gap can be observed between medical data (red points) and general natural images (blue/green points). 
} \label{fig1}
\end{figure}

In this paper, we try to answer this question by introducing a new simple yet effective medical segmentation pipeline, by only looking at a few exemplars, while still generalise well.
Specifically, the proposed approach only takes a few exemplar medical image pairs as input, and transfers the prior knowledge learned from general natural images to the target medical domain, with a new selective adaptation approach. The number of exemplars can be as few as one or five, and the prior knowledge is based on a large model (\eg SAM~\cite{kirillov2023segment}) pre-trained for natural image segmentation.

Directly applying this model to medical segmentation would fail in achieving similar performance as in natural images, due to the underlying domain gap (as illustrated in Fig.~\ref{fig1}). This further introduces the challenge of how to adapt the prior knowledge to the target medical domain.
To address this issue, we propose a domain-aware selective adaptation module, which enables lightweight training of large model to transfer the learned representations to the target domain, without fine-tuning the original large model parameters.

With the newly proposed learning pipeline, the model is able to easily adapt to different medical domains by applying the proposed transfer approach with only a few domain-specific exemplars. This is validated over a variety of medical image modalities, with significantly better performance than state-of-the-art methods under the same setting. To summarise, the main contributions of our work are as follows:
\begin{itemize}
    \item We propose, to our knowledge, the first attempt towards adapting general prior knowledge to various medical domains, by exposing only a few exemplars.
    \item We introduce a new domain-aware selective adaptation approach, which enables simple yet effective adaptation of large pre-trained models and boosts their performance in target domains.
    \item We identify the issues with the use of prompts in existing prompt-based medical image segmentation models, and propose a coarse prompt setting that better aligns with real-world scenarios.
    \item Extensive experiments validate the effectiveness of the proposed method, achieving state-of-the-art performance under the challenging few exemplar setting, surpassing existing works by a large margin.
\end{itemize}

\section{Related Work}

\subsubsection{Medical image segmentation} Medical Image Segmentation is essential for diagnostics and treatment planning \textcolor{black}{by} leveraging MRI, CT, and Ultrasound images to identify and delineate anatomical structures. The advent of Deep Learning (DL) and Convolutional Neural Networks (CNNs) has significantly advanced automated segmentation by enabling models to learn knowledge from large annotated datasets, thus improving accuracy. A significant innovation in the field of image segmentation is the Segment Anything model (SAM) ~\cite{kirillov2023segment}, which adopts a prompt-based approach similar to large language models such as Generative Pre-trained Transformers (GPT). SAM employs a Vision Transformer (ViT) backbone for encoding and a mask decoder for generating segmentation masks, \textcolor{black}{demonstrating} remarkable generalisation across diverse datasets with task-specific training.

However, such SAM-like models face challenges in medical image segmentation due to the scarcity of annotated datasets and inherent domain differences like variations in colour, texture, and anatomical structures. These challenges underscore the necessity for semi-supervised, unsupervised learning, and transfer learning strategies to effectively utilise unlabelled data and knowledge from related domains. Recent advancements focus on integrating domain-specific knowledge and employing techniques like attention mechanisms and generative adversarial networks to enhance model generalisability and produce realistic synthetic training data. Despite their success in natural image segmentation, adapting them to medical domains remains a challenge, driving ongoing research to overcome data scarcity, enhance model adaptability, and ensure clinical relevance and interpretability in medical image segmentation.

\subsubsection{Adapter-based transfer learning} In the field of machine learning, especially with large pre-trained models in natural language processing and computer vision, Parameter-Efficient Fine-Tuning (PEFT)~\cite{fu2023effectiveness} has emerged as a crucial technique for enhancing fine-tuning efficiency and effectiveness without substantial computational costs. Liu~\etal~\cite{liu2023explicit} introduce a novel adapter design that extracts frequency domain information as explicit visual prompts, integrating it into the intermediate layers of the SAM model. This enhances the model performance in low-level structure segmentation tasks. Similarly, Zhang~\etal~\cite{zhang2023lora} applied a low-rank adaptation (LoRA)~\cite{hu2021lora} fine-tuning strategy to the SAM image encoder and trained \textcolor{black}{the adapter parameters} alongside the SAM mask decoder to customise SAM for abdominal segmentation tasks. 

Adapter-based model tuning, a key method within PEFT, involves integrating lightweight adapter layers into existing models, allowing for task-specific adjustments through a minimal number of trainable parameters. This approach effectively leverages the intrinsic knowledge of pre-trained models while avoiding extensive retraining, thus preventing overfitting and facilitating rapid adaptation to new tasks with significantly reduced data requirements. Adapters offer a scalable solution for customising complex models across various applications, representing a key advancement in making large-scale machine learning models more accessible and adaptable for specialised tasks.

\subsubsection{Prompt-based segmentation} 
Segment Anything Model (SAM)~\cite{kirillov2023segment} is a pioneering work in prompt-based segmentation domain, being the first to develop a promptable segmentation model that has been pre-trained on a vast dataset. With appropriate prompts, SAM can generate possible masks for targets without specific task training, which can be more easily fine-tuned for downstream tasks. Most related work in the medical field has concentrated on fine-tuning SAM for specific segmentation datasets, particularly because SAM shows significant performance degradation on medical images segmentation. For instance, MedSAM~\cite{ma2024segment} fine-tuned SAM's decoder using prompts generated from label masks across over 30 medical image datasets. SAM-Med2D~\cite{doe2023sammed2d} enhances the SAM model by incorporating medical-specific prompts and fine-tuning on extensive medical image datasets, significantly improving its segmentation performance in medical imaging tasks. This fine-tuning led to improved performance compared to zero-shot predictions using the original SAM prompts. 

However, in practical \textcolor{black}{clinical}  scenarios, providing accurate prompts for a large volume of medical data can become cumbersome, especially when organs and tissues are small and adjacent to each other. Moreover, \textit{if the GT masks used to generate the prompts are already available, why bother segmentation to generate the masks} Creating precise prompts often requires domain-specific expertise, significantly increasing the training cost of the model. Above all, it becomes practically significant to have a model performing efficient inference without relying on precise prompts and extensive labelled training data during the training and inference phases. Such a model would align better with real-world applications, addressing the challenges of prompt accuracy and the availability of highly accurate labelled data.

\begin{figure} [t]
\begin{center}
    \includegraphics[width=1\textwidth]{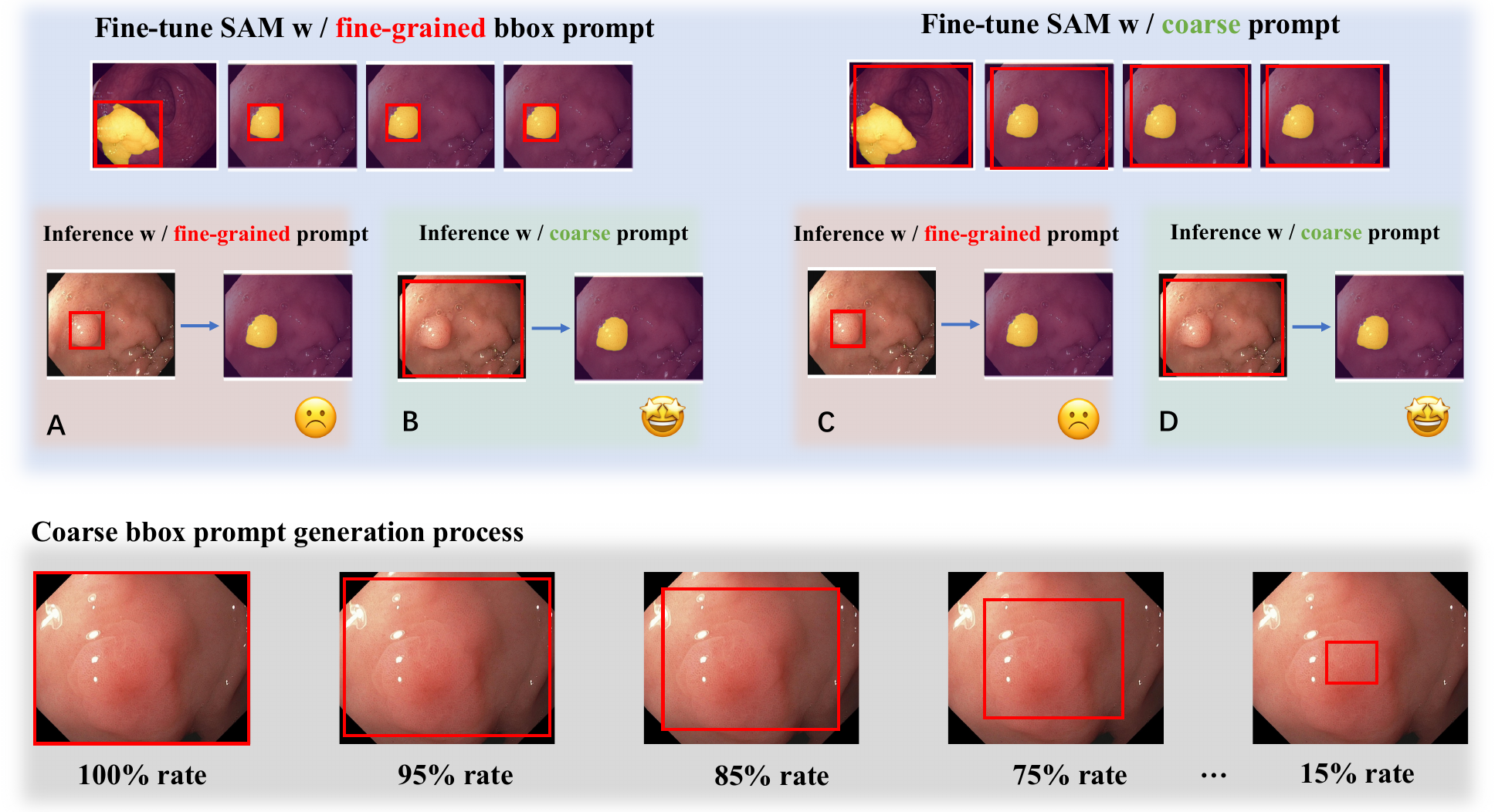}
\end{center}
\caption{Four settings of using bbox prompts during training and testing stages. The coarse bounding box prompt is designed to be GT-agnostic, with different ratios indicating the proportion of pixels by which the box region is shrunk inward relative to the entire image. Pseudo-code for coarse bbox prompt generation is shown in Algorithm \ref{alg:coarse}.}
\label{compa}
\end{figure}

\section{The Visual Prompt in Medical Image Segmentation}
The segment anything model (SAM) can adopt many types of visual prompts, \eg scribbles, clicks, or boxes to segment the arbitrary object within the image. It demonstrates highly generalised segmentation performance using prompts during training and testing. This paper focuses on the form of a bounding box prompt. Consequently, mainstream approaches to applying SAM for medical image segmentation follow the setting: utilising prompts in both training and testing. We argue that the prompts used in previous methods for medical segmentation is inappropriate \textcolor{black}{for clinical scenarios}. We categorise prompts into two types: \textbf{fine-grained prompts and coarse prompts}. The fine-grained prompts, as shown in Fig. \ref{compa} A and C, are customary user-provided or generated from manually annotated results. They are bespoke for each image and provide strong prior knowledge of the target location. Coarse prompts, as illustrated in Fig.~\ref{compa} B and D, remain consistent across different images and offer almost no prior knowledge. Note that our definitions of fine-grained and coarse prompts differ from those in~\cite{yang2024fine}. Based on these two types of prompts, as shown in Fig. \ref{compa}, there are four settings of using bbox prompts during training and testing: \textbf{Setting A}: trained with fine-grained bbox prompts and tested with fine-grained bbox prompts. \textbf{Setting B}: trained with fine-grained bbox prompts and tested with coarse bbox prompts. \textbf{Setting C}: trained with coarse bbox prompts and tested with fine-grained bbox prompts. \textbf{Setting D}: trained with coarse bbox prompts and tested with coarse bbox prompts.

\begin{algorithm}[t]
\caption{Pseudocode (PyTorch-style) for coarse bbox prompt generation}
\label{alg:coarse}
\definecolor{codeblue}{rgb}{0.25,0.5,0.5}
\lstset{
  backgroundcolor=\color{white},
  basicstyle=\fontsize{7.2pt}{7.2pt}\ttfamily\selectfont,
  columns=fullflexible,
  breaklines=true,
  captionpos=b,
  commentstyle=\fontsize{7.2pt}{7.2pt}\color{codeblue},
  keywordstyle=\fontsize{7.2pt}{7.2pt},
}
\begin{lstlisting}[language=python]
B, C, W, H = image.shape
# the image is a square size
offset = args.bbox_rate * W

# generate the coarse bbox with rate
box_prompt = torch.tensor([[[W - offset, H - offset, offset, offset]]], device=model.device, dtype=torch.float64)

# using the coarse bbox prompt
outputs = model(pixel_values = batch["pixel_values"].to(model.device), 
    input_boxes = box_prompt,
    multimask_output = False)

\end{lstlisting}
\end{algorithm}

Most SAM adapters in medical image segmentation rely on user-provided prompts or assume prompts generated from segmentation annotations, \ie the lesion area is already known, and a bounding box prompt for the lesion area is given, expecting the model to accurately segment the lesion within this region (setting A). However, this assumption is not applicable in real diagnostic scenarios. For unseen samples, the lesion area is unknown, making it infeasible to provide such precise fine-grained prompts. Therefore, a prompt setting that aligns more with real-world applications should be settings B and D, where only a coarse bounding box prompt is provided during inference, \eg a box region almost the same size as the original image. \textcolor{black}{Inconsistent setting of prompt used in training and testing may affect the performance.} Thus, this paper primarily investigates setting D in Fig. \ref{compa}. It is more challenging and practical compared to the other settings since there is no accurate lesion area information provided. 


\section{Method}
\begin{figure}[t]
\begin{center}
    \includegraphics[width=1.0\textwidth]{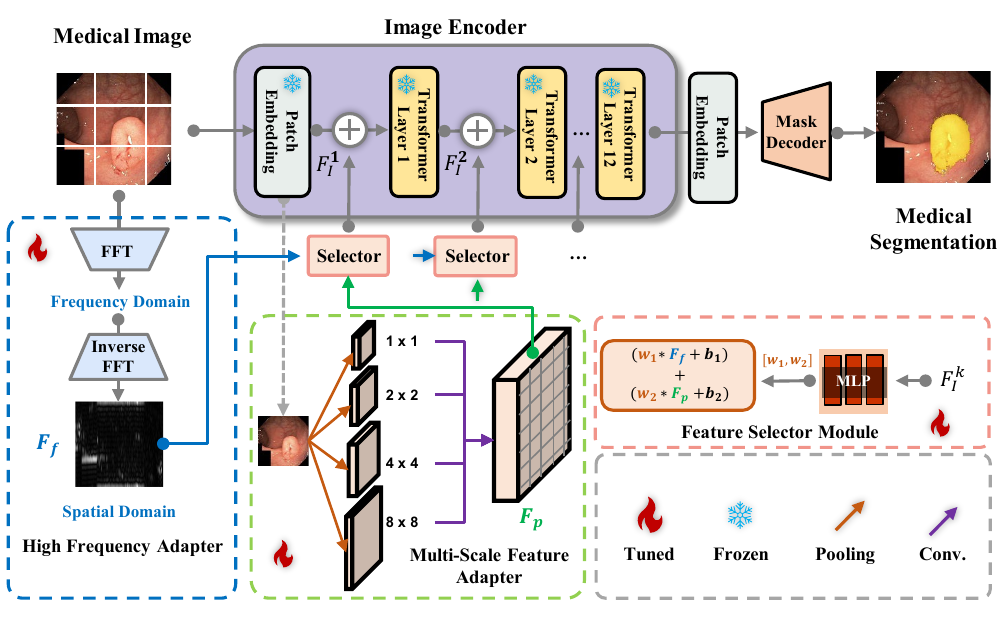}
\end{center}
\caption{\textcolor{black}{The proposed FEMed architecture. The pre-trained SAM image encoder is equipped with two specialised Adapters: (a) the Multi-Scale Features Adapter that captures features at various granularities through pyramid pooling, and (b) the High-Frequency Adapter that emphasises salient textural details via frequency domain analysis. The output features from these Adapters are fed into the Selection Module which contains a trainable decision layer that takes $F_k^I$ (where $k$ refers to the features from the $k$-th layer) as input to generate the weights for aggregating $F_f$ and $F_p$.
}} \label{fig:pipeline}
\end{figure}

In this paper, we propose an approach to better leverage and adapt general prior knowledge for medical image segmentation. Instead of fine-tuning all the network parameters, we retain the pre-trained image encoder parameters to leverage the priors from large-scale training. The overall model design is shown in Fig. \ref{fig:pipeline}. Specifically, our FEMed model comprises three key components: the High Frequency Adapter extracts frequency domain features via fast Fourier transform; a Multi-Scale Feature Adapter captures multi-scale features through adaptive pooling; the Feature Selector learns to determine which features to fuse with the intermediate results of each image encoder transformer block. By freezing all parameters of the image encoder and only training the adapters, the model efficiently learns medical knowledge with fewer training \textcolor{black}{exemplars}. Considering practical medical application scenarios, our method does not rely on fine-grained prompts for both training and testing phases. Alternatively, the design aims to accurately segment lesion areas under the coarse visual prompt setting.

\subsubsection{High frequency adapter}

We initially apply the Fast Fourier Transform (FFT) to the original image. The FFT effectively transforms the image from the spatial domain to the frequency domain, allowing us to isolate and analyse the high-frequency components. Following the FFT, we perform patch embedding as described in~\cite{liu2023explicit}, which involves segmenting the transformed image into smaller patches and embedding these patches into a suitable feature space. These resulting embeddings, rich in high-frequency information, are then combined with the original image embeddings. This combination ensures that both spatial and frequency domain features are retained and integrated. The combined embeddings are subsequently fed into a lightweight Multi-Layer Perceptron (MLP) with learnable parameters. The role of this MLP is to process the embeddings and produce clues that encapsulate additional insights from the frequency domain. The output clues generated by the MLP captures the nuanced details from the frequency domain and incorporates them into the image analysis process. To facilitate further operations, the clues is reshaped into a four-dimensional tensor. This reshaping can be viewed as an inverse transform, where the previously obtained clue is converted back to its original dimensional format, ready for subsequent processing stages. By focusing on high-frequency feature extraction, this approach equips the model with enhanced capabilities to discern subtle details within the image. Such detailed analysis leads to significant performance improvements, allowing for more accurate and reliable image analysis outcomes.

\subsubsection{Multi-scale features adapter}

The Multi-Scale Feature Adapter is designed to leverage the hierarchical feature of spatial data in medical images. By aggregating features across different scales, this adapter captures both global and local contexts, which are crucial for interpreting complex visual patterns, especially in medical images.
The process begins with the input feature map being subjected to a global average pooling layer. This layer consolidates the spatial information into a compact representation, effectively capturing the overall context of the input features. Following this, the condensed feature map undergoes channel processing, which aims to learn and enhance the correlations between different channels. Subsequently, the channel-processed feature map is fed through a pooling pyramid architecture, which consists of four adaptive average pooling layers, each producing output sizes of \{$1\times1$, $2\times2$, $4\times4$, $8\times8$\}, respectively. These varying scales of feature maps are critical as they encapsulate information from different levels of detail within the image. Once these multi-scale feature maps are generated, they are resized back to the original image dimensions using interpolation operations. This step ensures that all feature maps, despite their different scales, are aligned in size and can be effectively merged. The final step involves fusing these resized multi-scale feature maps into a single, comprehensive feature representation. This fusion process integrates the diverse information captured at various scales, providing a holistic view of the image features. The Multi-Scale Features Adapter is essential for extracting and merging multi-scale features from the input feature map, resulting in a robust new feature representation. This multi-scale approach ensures that the model is responsive to features of all sizes, significantly enhancing its ability to detect and recognise patterns across different spatial scales. This capability is particularly crucial for tasks such as medical image segmentation, where precision in identifying features of varying scales can greatly impact the accuracy and effectiveness of the analysis.

\subsubsection{Selection module}

And the core of our approach is the Feature Fusion Selector, a novel mechanism that dynamically determines the integration strategy for image features extracted by the Multi-Scale Features Adapter and the High Frequency Adapter. The selector employs a decision-making process based on learnable weights $\mathbf{W} = [w_1, w_2]$, \textcolor{black}{ where $\mathbf{W} = Softmax(MLP(F_I^k))$, $k$ refers to the features from
the k-th layer, $F_I$ is the feature embedding from each transformer layer. The aggregated image feature $\mathcal{F}$ is then updated as:}

\begin{equation}
    \textcolor{black}{\mathcal{F} = (w_1 * F_f + b_1) + (w_2 * F_p + b_2)}
\end{equation}
where \textcolor{black}{$F_f$ denotes the high-frequency feature, and $F_p$ represents the multi-scale feature. The bias term $\mathbf{B}=[b_1, b_2]$ is a learnable balancing factor that adjusts the importance between the multi-scale and high-frequency features. The weights $w_1$ and $w_2$ are learnable parameters constrained between [0,1], representing the relative importance of the features.}

The selector employs a linear layer to generate decision scores based on the input features, \textcolor{black}{which} are then passed through a softmax function to produce selection weights. The selection process determines which type of feature (frequency domain or multi-scale) to emphasise. This \textcolor{black}{feature} selection ensures clear physical significance and enhances the decision-making process of the model. The fusion of features is \textcolor{black}{aggregated} by dynamically adjusting the weights assigned to the frequency domain and multi-scale features. This \textcolor{black}
{structure} allows the model to dynamically adjust the contribution of each feature type at each transformer layer, enhancing the model capability to capture and integrate relevant features for medical image segmentation tasks. To address the observation that multi-scale features contribute more significantly to segmentation accuracy on medical tasks, a learnable bias term is introduced. This bias term increases the weight of the multi-scale features \textcolor{black}{by initialising bias terms to $b_1=0, b_2=1$ during the fusion process, the value of the weights will be updated as the model is optimised.}

\subsubsection{Loss function}
Dice loss is used due to its effectiveness in addressing class imbalance and handling small target regions as it directly measures the overlap between the predicted and actual segmentation. As a supplement, Cross-Entropy Loss, being a pixel-level loss function, is more \textcolor{black}{appropriate} for large target regions. \textcolor{black}{The combination of these two loss functions ensures} robust performance in medical image segmentation tasks. As a result, the loss function used for our model training is:
\begin{equation}
    \mathcal{L}=\alpha \mathcal{L}_{Dice} + \beta \mathcal{L}_{CE}.
\end{equation}

\section{Experiment}

\subsection{Implementation Details}
Our model is implemented using PyTorch and MONAI~\cite{cardoso2022monai} and trained on a single NVIDIA 4090 GPU. The learning rate is set to 0.0001 with a decay of 0.01. All input images are normalised to have a zero mean and unit standard deviation for non-zero voxels. During training, 3D images and labels are sliced using a sliding window approach with a two-dimensional size of $256\times256$. Additionally, we perform intensity scaling on the images to adjust the pixel intensity values from a range of -1,000 to 2,000 to a standardised range of 0 to 255, and similarly, label intensities are adjusted within the same output range for consistency and normalisation purposes. The batch size is set to 1. The number of training \textcolor{black}{exemplars} ranges within \{$1,5,10$\}. The default coarse bbox rate is 0.95. Unless otherwise specified, the default setting of using bbox during training and testing is D. Following~\cite{ma2024segment}, we resize all the images to a larger scale with $1,024\times 1,024$ to achieve better performance.

All models are trained for a total of 100 epochs using the Adam optimiser. Our experiment employs an early stopping mechanism with 10 epochs tolerance to prevent overfitting during model training. The decay rate is applied every 30 epochs. The balancing factors $\alpha$ and $\beta$ in the loss function are both set to 1.

\subsection{Datasets}

Our study leverages multiple publicly available benchmarks to validate the effectiveness of the proposed segmentation approach across three different medical imaging \textcolor{black}{datasets}. Each dataset comes with its unique challenges and characteristics, catering to the segmentation of various anatomical structures and pathologies. Here, we describe the datasets used and our approach to validation and testing. The Brain Tumor Segmentation (BraTS) 2021 challenge dataset~\cite{baid2021rsna} is widely used for medical segmentation methods. It includes MRI scans of 1,251 subjects across four 3D MRI modalities. Liver Tumor Segmentation (LiTS) 2017~\cite{umer2020maccai} challenge dataset. This dataset includes CT scans for the task of liver tumour and liver segmentation, presenting a diverse set of imaging conditions and tumour appearances. We also employed the RGB colour image dataset Kvasir-Seg~\cite{jha2020kvasir}, specifically designed for pixel-level segmentation of colorectal polyps. This dataset comprises 1,000 images of gastrointestinal polyps along with their corresponding segmentation masks, all meticulously annotated and verified by expert gastroenterologists. Our experiments on this dataset validate the efficacy of our method in handling colour medical images. 

\begin{table}[t]\centering
\caption{Comparison of our method with SAM and SOTA segmentation methods, without fine-grained bbox prompt on multi-modal medical datasets. \textcolor{black}{SAM~\cite{kirillov2023segment}, MedSAM~\cite{ma2024segment} and SAM-MED2D~\cite{cheng2023sam} are included in comparison.} The $\star$ represents setting A, while $\dagger$ indicates setting D. NTS: number of training \textcolor{black}{exemplars}. If NTS is not specified, it indicates that we directly loaded the pre-trained weights provided by the original methods and tested the results without any additional training.}
\label{main_res}
\begin{tabular}{l|c|ccc|ccc|ccc}
\toprule
                         & \multicolumn{1}{l|}{}                      & \multicolumn{3}{c|}{\begin{tabular}[c]{@{}c@{}}BraTS21\\ MRI\end{tabular}}                                               & \multicolumn{3}{c|}{\begin{tabular}[c]{@{}c@{}}LiTS17\\ CT\end{tabular}}                                                  & \multicolumn{3}{c}{\begin{tabular}[c]{@{}c@{}}Kvasir-SEG\\ RGB\end{tabular}}                                              \\ \cline{3-11} 
\multirow{-2}{*}{Method} & \multicolumn{1}{l|}{\multirow{-2}{*}{NTS}} & \multicolumn{1}{l}{Dice$\uparrow$}     & \multicolumn{1}{l}{Hd95$\downarrow$}   & \multicolumn{1}{l|}{mIoU$\uparrow$}    & \multicolumn{1}{l}{Dice$\uparrow$}     & \multicolumn{1}{l}{Hd95$\downarrow$}    & \multicolumn{1}{l|}{mIoU$\uparrow$}    & \multicolumn{1}{l}{Dice$\uparrow$}     & \multicolumn{1}{l}{Hd95$\downarrow$}    & \multicolumn{1}{l}{mIoU$\uparrow$}     \\ \midrule
\rowcolor[HTML]{EFEFEF} 
SAM $\star$              & -                                          & 55.31                                  & 19.36                                  & 42.27                                  & 59.76                                  & 48.46                                   & 47.21                                  & 74.04                                  & 66.96                                   & 61.35                                  \\
\rowcolor[HTML]{EFEFEF} 
MedSAM $\star$           & -                                          & 58.20                                  & 16.23                                  & 44.18                                  & 48.84                                  & 45.94                                   & 35.63                                  & 81.81                                  & 35.16                                   & 70.83                                  \\
\rowcolor[HTML]{EFEFEF} 
SAM-Med2D $\star$        & -                                          & 60.01                                  & 14.01                                  & 46.24                                  & 52.23                                  & 46.23                                   & 38.23                                  & 79.92                                  & 45.09                                   & 67.88                                  \\ \hline
\rowcolor[HTML]{EFEFEF} 
SAM $\dagger$            & -                                          & 8.80                                   & 141.55                                 & 4.79                                   & 29.85                                  & 169.41                                  & 19.95                                  & 48.18                                  & 133.72                                  & 36.93                                  \\
\rowcolor[HTML]{EFEFEF} 
MedSAM $\dagger$         & -                                          & 4.32                                   & 135.18                                 & 2.26                                   & 13.64                                  & 146.18                                  & 8.40                                   & 48.04                                  & 133.98                                  & 36.74                                  \\
\rowcolor[HTML]{EFEFEF} 
SAM-Med2D $\dagger$      & -                                          & 4.28                                   & 72.46                                  & 2.88                                   & 14.32                                  & 143.21                                  & 9.92                                   & 48.32                                  & 133.55                                  & 36.09                                  \\ \hline
SAM$\dagger$             &                                            & 12.79                                  & 52.52                                  & 9.06                                   & 40.04                                  & 116.28                                  & 28.09                                  & 25.17                                  & 104.63                                  & 16.52                                  \\
MedSAM$\dagger$          &                                            & 12.78                                  & 51.50                                  & 8.84                                   & 12.07                                  & 114.01                                  & 7.79                                   & 31.43                                  & 105.96                                  & 17.50                                  \\
SAM-Med2D$\dagger$       & \multirow{-2}{*}{1}                        & 12.20                                  & 51.56                                  & 9.02                                   & 15.23                                  & 113.32                                  & 10.21                                  & 7.77                                   & 132.21                                  & 4.37                                   \\
Ours $\dagger$           & \multicolumn{1}{l|}{}                      & \cellcolor[HTML]{ECF4FF}\textbf{14.43} & \cellcolor[HTML]{ECF4FF}\textbf{48.55} & \cellcolor[HTML]{ECF4FF}\textbf{9.66}  & \cellcolor[HTML]{ECF4FF}\textbf{42.60} & \cellcolor[HTML]{ECF4FF}\textbf{108.08} & \cellcolor[HTML]{ECF4FF}\textbf{29.72} & \cellcolor[HTML]{ECF4FF}\textbf{28.83} & \cellcolor[HTML]{ECF4FF}\textbf{100.05} & \cellcolor[HTML]{ECF4FF}\textbf{18.24} \\ \hline
SAM$\dagger$             &                                            & 29.84                                  & 49.95                                  & 22.32                                  & 51.30                                  & 82.72                                   & 39.87                                  & 43.04                                  & 102.18                                  & 31.00                                  \\
MedSAM$\dagger$          &                                            & 29.42                                  & \textbf{41.73}                         & 21.41                                  & 47.18                                  & 84.79                                   & 35.34                                  & 39.13                                  & 103.60                                  & 27.16                                  \\
SAM-Med2D$\dagger$       & \multirow{-2}{*}{5}                        & 27.79                                  & 43.80                                  & 24.83                                  & 52.56                                  & 79.78                                   & 41.23                                  & 40.08                                  & 106.79                                  & 28.18                                  \\
Ours $\dagger$           &                                            & \cellcolor[HTML]{ECF4FF}\textbf{34.12} & \cellcolor[HTML]{ECF4FF}43.50          & \cellcolor[HTML]{ECF4FF}\textbf{25.19} & \cellcolor[HTML]{ECF4FF}\textbf{55.98} & \cellcolor[HTML]{ECF4FF}\textbf{77.68}  & \cellcolor[HTML]{ECF4FF}\textbf{44.16} & \cellcolor[HTML]{ECF4FF}\textbf{47.80} & \cellcolor[HTML]{ECF4FF}\textbf{99.28}  & \cellcolor[HTML]{ECF4FF}\textbf{35.13} \\ \hline
SAM$\dagger$             &                                            & 34.08                                  & 42.20                                  & 25.74                                  & 56.35                                  & 70.51                                   & 45.03                                  & 52.12                                  & 114.54                                  & \textbf{40.47}                         \\
MedSAM$\dagger$          &                                            & 37.72                                  & 43.79                                  & 28.67                                  & 52.17                                  & 73.17                                   & 41.96                                  & 47.56                                  & 116.64                                  & 35.06                                  \\
SAM-Med2D$\dagger$       & \multirow{-2}{*}{10}                       & 41.60                                  & 37.29                                  & 31.04                                  & 56.95                                  & 68.21                                   & 43.23                                  & 49.93                                  & 121.49                                  & 37.67                                  \\
Ours $\dagger$           &                                            & \cellcolor[HTML]{ECF4FF}\textbf{51.29} & \cellcolor[HTML]{ECF4FF}\textbf{32.45} & \cellcolor[HTML]{ECF4FF}\textbf{39.70} & \cellcolor[HTML]{ECF4FF}\textbf{58.20} & \cellcolor[HTML]{ECF4FF}\textbf{65.96}  & \cellcolor[HTML]{ECF4FF}\textbf{48.26} & \cellcolor[HTML]{ECF4FF}\textbf{52.34} & \cellcolor[HTML]{ECF4FF}\textbf{113.36} & \cellcolor[HTML]{ECF4FF}39.80          \\ \bottomrule
\end{tabular}
\end{table}
\subsection{Quantitative and Qualitative Analysis}
The quantitative results of our proposed method are shown in Table \ref{main_res}. Under the few-shot setting, where only 1, 5, or 10 \textcolor{black}{exemplars} are used for training, our method consistently achieves the best performance. Specifically, compared to previous models, our method shows significant improvements on various medical modalities/datasets. On the BraTS21 dataset, our method achieves a performance gain of 0.6, 2.87, and 8.66 mIoU for the respective exemplar sizes. On the LiTS17 dataset, the mIoU improvements are 1.63, 2.93, and 3.23. On the Kvasir-SEG dataset, our approach also improves the 1- and 5-exemplar settings by 0.74 and 4.13 in mIoU.

Moreover, changing the bbox prompt setting from A to B (\ie more practical test setting) led to a significant performance drop for \textcolor{black}{SAM~\cite{cheng2023sam}, as well as for models fine-tuned on large medical datasets such as MedSAM~\cite{ma2024segment} and SAM-MED2D~\cite{cheng2023sam}} (as shown in the shaded first six rows of Table \ref{main_res}). This drop is reasonable due to the inconsistency between training and inference settings. However, when retrained with consistent training and testing settings (\ie setting D), the performance of these models was far worse than under setting A. This surprising result suggests that models performing exceptionally well under controlled settings (A) may not generalise effectively to real-world scenarios (D). This insight indicates that existing studies in the literature on SAM-adapter for medical image segmentation are arguably not on the right track, and should be rectified. Furthermore, under setting D, SAM overall outperforms MedSAM and SAM-Med2D. This interesting observation suggests that in real-world medical applications with a limited number of training \textcolor{black}{exemplars}, the original SAM already shows better generalisability and applicability compared to models further trained on large medical datasets with paired GT mask labels.

\begin{figure}
\begin{center}
    \includegraphics[width=1\textwidth]{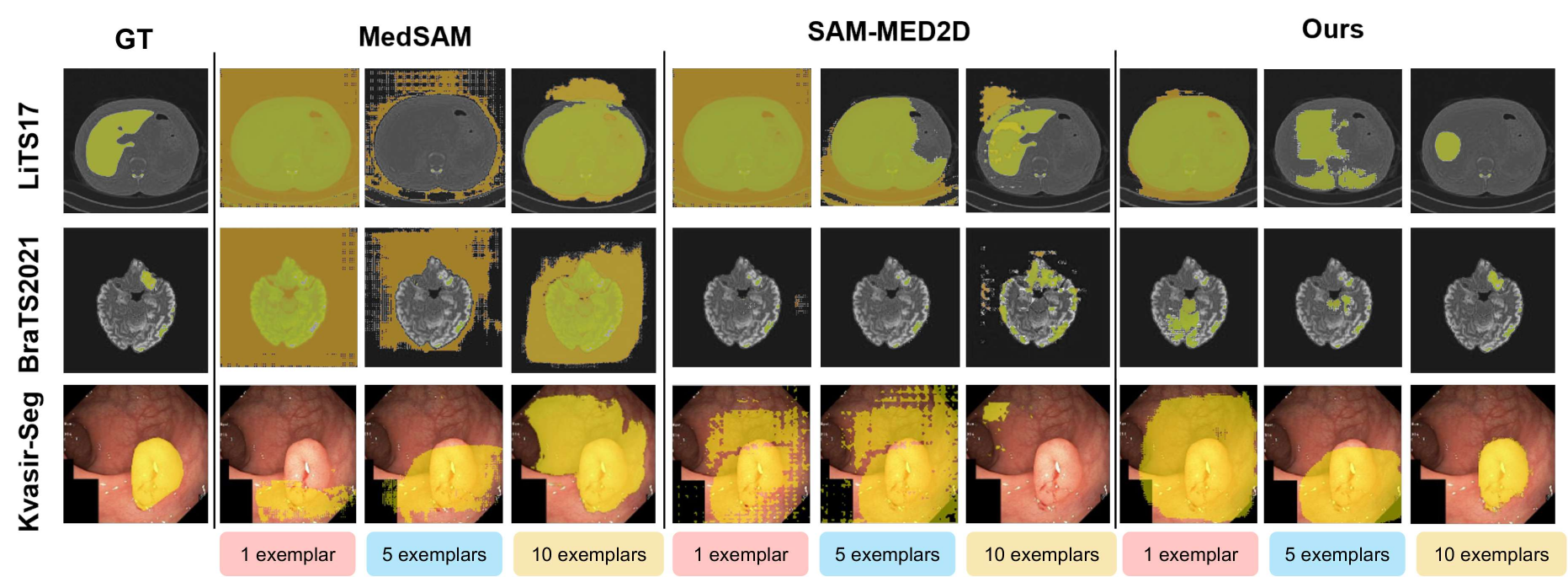}
\end{center}
\caption{\textcolor{black}{Qualitative performance across three medical datasets (LiTS17, BraTS2021, and Kvasir-Seg) using different methods: MedSAM~\cite{ma2024segment}, SAM-MED2D~\cite{cheng2023sam}, and our proposed method (``Ours''). For each method, we show the segmentation results with different numbers of \textcolor{black}{exemplars} (\ie 1, 5, and 10).}}
\label{quali}
\end{figure}

Qualitative results in Fig. \ref{quali} further illustrate our findings. When trained and tested using setting D, which employs a coarse bbox prompt covering 95\% of the original image, predictions from previous methods tend to almost cover the entire image, failing to locate the lesion area. In contrast, our method can well segment the lesion area, even for very small regions, as demonstrated by the qualitative results shown in the second row of Fig. \ref{quali}. This precision highlights the robustness and accuracy of our approach, even under challenging conditions. The ability of our method to accurately identify and segment lesion areas, despite the coarse bounding box prompts, underscores its potential for practical applications in medical image segmentation.

\subsection{Ablation study}

\subsubsection{Effectiveness of the proposed components} 
We conducted ablation experiments to analyse the different components of our proposed method. The results from Table \ref{abla_me} illustrate the effectiveness of each component. Particularly, the High-Frequency Adapter (HFA) module contributes more compared to the Multi-Scale Features Adapter (MSFA) module under setting D. 

\begin{table}[]\centering
\caption{Ablation study on each component. Results are tested by training models with 10 \textcolor{black}{exemplars}. HFA: High Frequency Adapter, MSFA: Multi-Scale Features Adapter.}
\label{abla_me}
\begin{tabular}{ccc|ccc|ccc|ccc}
\toprule
\multicolumn{3}{c|}{Component}       & \multicolumn{3}{c|}{BraTS}                        & \multicolumn{3}{c|}{LiTS17}                                                                   & \multicolumn{3}{c}{Kvasir-SEG}                    \\ \hline
HFA       & MSFA      & Selector  & Dice$\uparrow$ & Hd95$\downarrow$ & IoU$\uparrow$ & Dice$\uparrow$                & Hd95$\downarrow$              & IoU$\uparrow$                 & Dice$\uparrow$ & Hd95$\downarrow$ & IoU$\uparrow$ \\ \midrule
\ding{52} & \ding{55} & \ding{55} & 44.56          & 38.21            & 35.33         & 56.31                         & 68.32                         & 46.57                         & 51.06          & 118.36           & 38.90         \\
\ding{55} & \ding{52} & \ding{55} & 42.19          & 49.23            & 31.36         & 49.21                         & 78.21                         & 39.08                         & 49.29          & 107.82           & 36.74         \\
\ding{52} & \ding{52} & \ding{55} & 48.23          & 35.34            & 37.45         & 55.21                         & 70.01                         & 45.24                         & 49.86          & 108.01           & 37.37         \\ \hline
\rowcolor[HTML]{ECF4FF} 
\ding{52} & \ding{52} & \ding{52} & 51.29          & 32.45            & 39.70         & \cellcolor[HTML]{ECF4FF}58.20 & \cellcolor[HTML]{ECF4FF}65.96 & \cellcolor[HTML]{ECF4FF}48.26 & 52.34          & 103.36           & 39.80         \\ \bottomrule
\end{tabular}
\end{table}

We hypothesise that the superior performance of the HFA module is due to its ability to capture high-frequency information, such as edges, textures, and fine details. This high-frequency information is crucial for accurate segmentation, particularly when using coarse bounding box (bbox) prompts. By leveraging these detailed features, the model can more precisely delineate the target areas, even with less precise initial prompts.

These findings highlight the importance of incorporating high-frequency feature extraction in our approach, significantly enhancing segmentation performance for medical images. This improvement is especially evident when dealing with coarse bbox prompts, which are more practical and aligned with \textcolor{black}{the scenarios of} real-world medical imaging applications.

\subsubsection{Effectiveness of learnable bias in the proposed selector module} 
We conducted an ablation study on the learnable bias term in the proposed selector module, as detailed in Table \ref{abla_bias}. In the absence of the learnable bias term, the high-frequency features and multi-scale features are directly combined using weighted fusion. The results indicate a significant improvement in the segmentation performance upon \textcolor{black}{applying} the learnable bias term.

\begin{table}[!h]\centering
\caption{Ablation study on learnable bias in the selector. The symbol \ding{55} indicates without adding learnable bias while symbol \ding{52} indicates with adding learnable bias.}
\label{abla_bias}
\begin{tabular}{cccc|ccc|ccc}
\toprule
\multirow{2}{*}{Bias}          & \multicolumn{3}{c|}{BraTS21}                                                                                    & \multicolumn{3}{c|}{LiTS17}                                                                                     & \multicolumn{3}{c}{Kvasir-SEG}                                                                                 \\ \cline{2-10} 
                               & \multicolumn{1}{l}{Dice$\uparrow$} & \multicolumn{1}{l}{Hd95$\downarrow$} & \multicolumn{1}{l|}{mIoU$\uparrow$} & \multicolumn{1}{l}{Dice$\uparrow$} & \multicolumn{1}{l}{Hd95$\downarrow$} & \multicolumn{1}{l|}{mIoU$\uparrow$} & \multicolumn{1}{l}{Dice$\uparrow$} & \multicolumn{1}{l}{Hd95$\downarrow$} & \multicolumn{1}{l}{mIoU$\uparrow$} \\ \midrule
\multicolumn{1}{c|}{\ding{55}} & 42.01                              & 98.63                                & 29.44                               & 46.04                              & 81.50                                & 33.60                               & 24.43                              & 62.93                                & 15.89                              \\
\rowcolor[HTML]{ECF4FF}
\multicolumn{1}{c|}{\ding{52}} & 43.96                              & 99.46                                & 31.28                               & 46.85                              & 77.66                                & 35.10                               & 32.50                              & 52.77                                & 22.65                              \\ \bottomrule
\end{tabular}
\end{table}

This improvement suggests that the learnable bias term plays a crucial role in enhancing the fusion process by effectively modulating the contribution of each feature type. By allowing the model to adjust the bias dynamically, the integration of high-frequency and multi-scale features becomes more refined and accurate, leading to better overall performance. This finding underscores the importance of the learnable bias term in optimising the feature fusion mechanism within our segmentation framework.

\subsubsection{Impact of different bbox rates} 
\begin{figure}[!t]
\begin{center}
    \includegraphics[width=1.0\textwidth]{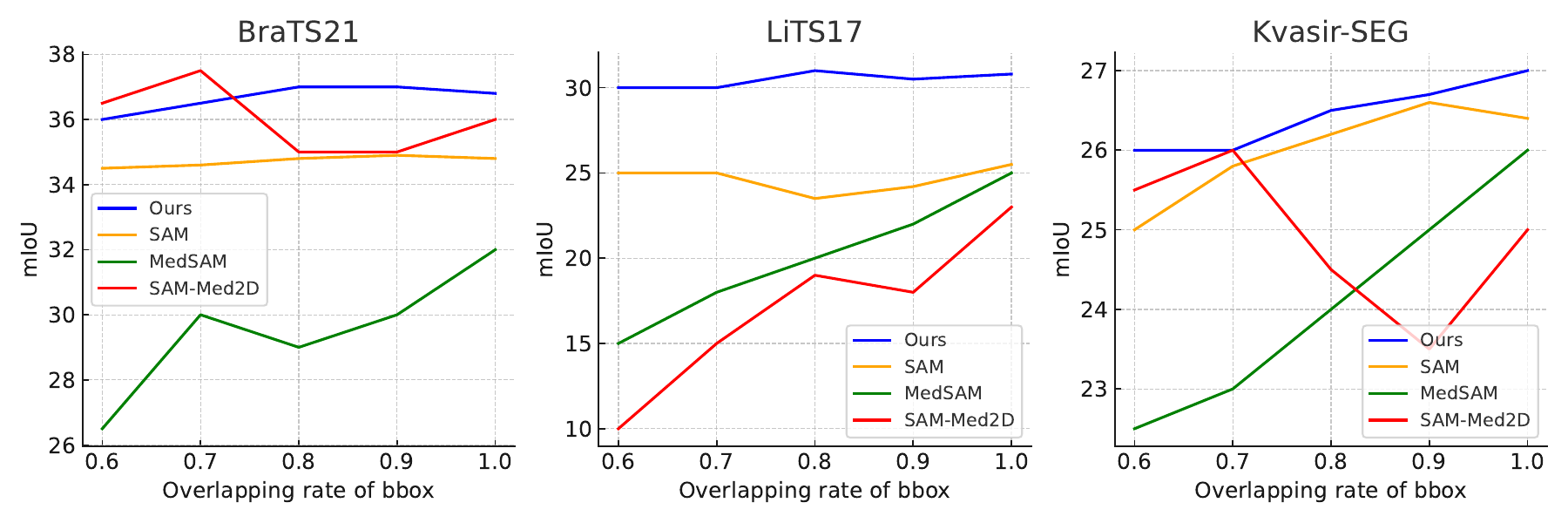}
\end{center}
\caption{The effect of varying bounding box overlapping rates \textcolor{black}{(refers to the proportion of pixels by which the box region is shrunk inward relative to the entire image, \ie the \textit{rate} in Fig.~\ref{compa}).} All results are reported via training with a single exemplar.} \label{rates}
\end{figure}

As shown in Fig. \ref{compa}, we can generate coarse prompts at different rates. To evaluate the impact of these variations, we conducted ablation experiments on coarse prompts at various rates, and the results are presented in Fig. \ref{rates}. The performance trends of different models with coarse bounding box (bbox) prompts at different rates show inconsistency. Overall, there is a tendency for performance to decrease as the rate decreases. This trend makes sense intuitively as a smaller rate tends to result in a bbox failing to fully encompass the lesion area, thus leading to poor performance.

Notably, in comparison to MedSAM~\cite{ma2024segment} and SAM-Med2D~\cite{doe2023sammed2d}, our method and SAM~\cite{kirillov2023segment} demonstrate relatively stable trends across different rates. This stability indicates that our approach, along with SAM, maintains robustness despite changes in the granularity of the prompts, ensuring more reliable segmentation outcomes under varying conditions. The observed stability in performance underscores the effectiveness of our method in maintaining high-quality segmentation even when the prompt rate varies, highlighting its potential for practical application in diverse and dynamic medical imaging scenarios.

\section{Conclusion}
In this paper, we presented a new approach for medical image segmentation that leverages only a few (\ie 5) exemplar inputs and domain-agnostic prior knowledge for adaptation to specific medical domains. Through selective adaptation, our method circumvents the need for extensive domain-expert annotations, addressing the challenge of scarce and costly medical segmentation data. Our extensive experiments across various medical image modalities demonstrate the effectiveness and superiority of our approach, marking a prominent advancement over current state-of-the-art methods. This not only validates the feasibility of achieving high-quality medical image segmentation with limited examples but also potentially \textcolor{black}{pave the way} for leveraging pre-trained models in medical imaging applications, thereby enhancing the potential for automation and efficiency in healthcare diagnostics. Additionally, we highlight the impracticality of previous prompt-based segmentation models that assume the availability of precise location of lesion areas for fine-grained bbox prompts, even during the inference phase. Based on this, we propose to use the coarse bbox prompts, which align better with real-world applications.

\subsubsection{\ackname}
This project is funded by the Royal Society International Exchanges Award (IES\textbackslash R3\textbackslash223050). J. Jiao is supported by the Royal Society Short Industry Fellowship (SIF\textbackslash R1\textbackslash231009) and the Amazon Research Award.


%
%
\bibliographystyle{splncs04}
\bibliography{main}
\end{document}